\pdfoutput=1

\documentclass[11pt]{article}
\PassOptionsToPackage{table,dvipsnames}{xcolor}
\usepackage{acl}

\usepackage{times}
\usepackage{latexsym}

\usepackage[T1]{fontenc}

\usepackage[utf8]{inputenc}

\usepackage{microtype}

\usepackage{longtable}
\usepackage{inconsolata}
\usepackage{hyperref}       
\usepackage{url}            
\usepackage{amsfonts}       
\usepackage{nicefrac}       
\usepackage[table]{xcolor}         
\usepackage{graphicx}
\usepackage{subfigure}
\usepackage{comment}
\usepackage{arydshln}
\usepackage{multirow}
\usepackage{wrapfig, lipsum}
\usepackage{enumerate}
\usepackage{enumitem}
\usepackage{tabularx}
\usepackage{balance}

\usepackage{algorithm}
\usepackage{algorithmic}

\usepackage{amsmath,amsfonts,bm}

\def\eqref#1{equation~\ref{#1}}

\def\1{\bm{1}}

\newcommand{\ie}{{\em i.e.,}}
\newcommand{\eg}{{\em e.g.,}}
\newcommand{\Ni}{({\em i})~}
\newcommand{\Nii}{({\em ii})~}
\newcommand{\Niii}{({\em iii})~}

\DeclareMathAlphabet{\mathsfit}{\encodingdefault}{\sfdefault}{m}{sl}
\SetMathAlphabet{\mathsfit}{bold}{\encodingdefault}{\sfdefault}{bx}{n}

\def\gL{{\mathcal{L}}}

\def\gV{{\mathcal{V}}}

\newcommand{\KL}{D_{\mathrm{KL}}}

\usepackage{amsmath}
\usepackage{amssymb}
\usepackage{mathtools}
\usepackage{amsthm}
\usepackage{caption}
\usepackage{booktabs}       
\usepackage{supertabular}
\usepackage{pifont}
\usepackage{fdsymbol}

\usepackage[capitalize,noabbrev]{cleveref}

\usepackage[textsize=tiny]{todonotes}
\newcommand{\ba}{BiAlign}
\definecolor{applegreen}{rgb}{0.56, 0.8, 0.25}
 
\usepackage{pgfplots}
\usetikzlibrary{pgfplots.groupplots}
\pgfplotsset{compat=1.3}
\usepackage{pifont}
\usepackage{microtype}

\newcommand{\okmark}{{\textbf{\textcolor[rgb]{0.1, 0.5, 0.1}{$\checkmark$}}}}
\newcommand{\ngmark}{{\textbf{\color{red}{\ding{55}}}}}
\usepackage{etoolbox}
\newcommand{\circled}[2][]{\tikz[baseline=(char.base)]
    {\node[shape = circle, draw, inner sep = 1pt]
    (char) {\phantom{\ifblank{#1}{#2}{#1}}};%
    \node at (char.center) {\makebox[0pt][c]{#2}};}}
\robustify{\circled}
\newcommand*{\Scale}[2][4]{\scalebox{#1}{$#2$}}%

\definecolor{extremelightgray}{gray}{0.9}

\title{Beyond Output Matching: Bidirectional Alignment \\ for Enhanced In-Context Learning}

\author{Chengwei Qin\textsuperscript{\ding{70}}$^\dagger$\thanks{\; Equal contribution, order decided by coin flip.}, Wenhan Xia$^\clubsuit$\footnotemark[1], Fangkai Jiao$^\dagger$, Chen Chen$^\dagger$, Yuchen Hu$^\dagger$, \\\textbf{Bosheng Ding$^\dagger$}, \textbf{Ruirui Chen$^\vardiamondsuit$}, \textbf{Shafiq Joty$^\dagger$$^\spadesuit$}\\
\textsuperscript{\ding{70}}The Hong Kong University of Science and Technology (Guangzhou) $^\clubsuit$Princeton University\\
$^\dagger$Nanyang Technological University $^\spadesuit$Salesforce Research\\
$^\vardiamondsuit$Institute of High Performance Computing (IHPC),\\
Agency for Science, Technology and Research (A*STAR), Singapore
}

\begin{document}
\maketitle
\begin{abstract}
Large language models (LLMs) have shown impressive few-shot generalization on many tasks via in-context learning (ICL). Despite their success in showing such emergent abilities, the scale and complexity of larger models also lead to unprecedentedly high computational demands and deployment challenges. In reaction, researchers explore transferring the powerful capabilities of larger models to more efficient and compact models by typically aligning the \emph{output} of smaller {(student)} models with that of larger {(teacher)} models. Existing methods either train student models on the generated outputs of teacher models or imitate their token-level probability distributions. However, these distillation methods pay little to no attention to the \emph{input}, which also plays a crucial role in ICL. Based on the finding that the performance of ICL is highly sensitive to the selection of demonstration examples, we propose Bidirectional Alignment (\ba) to fully leverage the models' preferences for ICL examples to improve the ICL abilities of student models. Specifically, we introduce the alignment of input preferences between student and teacher models by incorporating a novel ranking loss, in addition to aligning the token-level output distribution. With extensive experiments and analysis, we demonstrate that \ba\ can consistently outperform existing baselines on various tasks involving language understanding, reasoning, and coding.  
\end{abstract}

\section{Introduction} \label{sec:intro_bialign}

With the recent advancements in model scale and pretraining data, large language models (LLMs) have demonstrated impressive few-shot learning capabilities via in-context learning (ICL). With ICL, the LLM generates an output for a given query by conditioning on a few demonstration examples and optionally a task description, and it does so without any parameter updates \citep{brown2020language}. Despite the success of ICL in few-shot generalization, the high computational demands and deployment challenges posed by the size of the LLMs hinder their widespread application. Serving an LLM with 175B parameters requires
at least 350GB GPU memory \citep{hsieh-etal-2023-distilling}, which is far beyond what is affordable in most real-world settings. Also, the serving cost increases with model size -- it costs 1-2 FLOPs per parameter to infer on one token \citep{kaplan2020scaling}.

To alleviate this issue, researchers have proposed a number of methods to transfer the emergent capabilities of larger (teacher) models to more efficient and compact smaller (student) models, an approach commonly known as knowledge distillation \citep{hinton2015distilling}. In this approach, the student models are trained to align their \emph{output} space with that of the teachers. This is typically achieved by either training on the generated outputs of the teacher models \citep{hsieh-etal-2023-distilling,wang2022self,xu2023wizardlm} or by imitating their token-level probability distributions \citep{agarwal2023gkd,huang2022context,gu2024minillm}.\footnote{Different from the conventional \emph{strong-to-weak} generalization, \citet{burns2023weak} recently 
introduce \emph{weak-to-strong} generalization, which explores leveraging weaker (smaller) models to elicit ``superalignment'' from the stronger (larger) models. {This paper however considers the conventional \emph{strong-to-weak} approach.}}

\begin{figure*}[t]
  \begin{center}
   \includegraphics[width=0.96\textwidth]{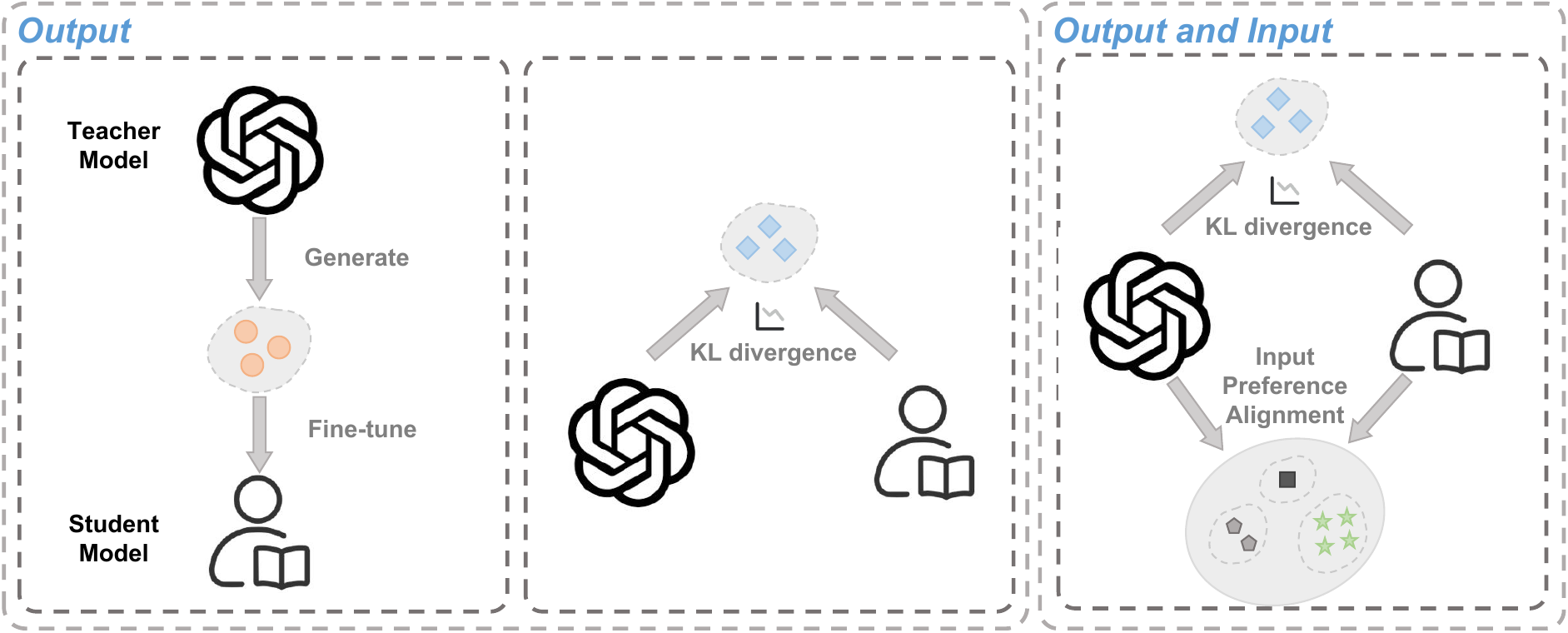}
  \end{center}
  \caption{Comparison between different types of approaches to aligning student models. Existing methods typically fine-tune student models on generated outputs of teacher models or to match their token-level output probability distributions (\emph{left} part). In contrast, our method (\ba) considers the models' preferences for different inputs {(the more helpful an input is for generating the target, the more the model prefers that input)} to achieve input preference alignment (\emph{right} part).}
  \label{fig:diff}
\end{figure*}

While existing distillation methods demonstrate improved ICL results, they pay little attention to the \emph{input}, specifically the demonstrations, which have been shown to have a significant impact on the performance of ICL \citep{zhao21c,xie2022an,qin2023context}. Indeed, selecting different sets of demonstration examples can yield performance ranging from almost random to better than state-of-the-art fine-tuned models \citep{gao-etal-2021-making,lu-etal-2022-fantastically}, indicating that the model has different preferences for different inputs. Inspired by this finding, we propose \textbf{Bidirectional Alignment} (\ba), a simple yet effective framework for improving the ICL abilities of student models (Figure~\ref{fig:diff}). Specifically, \ba\ introduces the alignment of input preferences between student and teacher models through the incorporation of a novel ranking loss, in addition to aligning the token-level output distributions. Our main hypothesis is that for an effective knowledge distillation, the student model should align with not only the teacher model's output distribution but also its input preference (i.e., the more helpful an input is for generating the target, the more the model prefers that input).\footnote{Our hypothesis is closely related to preference learning in RLHF, where the reward model learns `which outputs should be preferred'. After learning, a well-trained reward model can rank model responses with expertise comparable to humans.} \ba\ allows student models to obtain more fine-grained supervision from teacher models by fully leveraging their preferences for different demonstrations in ICL. Empirical results on tasks spanning language understanding, symbolic reasoning, mathematical reasoning, logical reasoning, and coding show that \ba\ can consistently outperform previous baselines. In summary, our main contributions are: 

\begin{itemize}[leftmargin=*]
    \item To the best of our knowledge, we for the first time consider aligning student models with teacher models from an \emph{input preference} perspective. We propose Bidirectional Alignment (\ba) to fully leverage the models' preferences for different demonstration examples to improve the ICL capabilities of student models.
    \item With extensive experiments and analysis, we demonstrate the effectiveness of \ba\ on a variety of tasks. For example, it brings about 20$\%$ relative improvement on GSM8K \citep{cobbe2021training} and 18$\%$ on LogiQA \citep{liu2020logiqa}.
\end{itemize}

\section{Related Work}

This work concerns how to improve the ICL ability of student models by aligning the student and teacher models' preferences for different few-shot demonstrations. In light of this, we review three lines of work that form the basis of this work: few-shot learning, in-context learning, and alignment.

\subsection{Few-shot Learning}

Few-shot learning (FSL) aims to learn tasks with only a few labeled examples, which faces the challenge of over-fitting due to the scarcity of labeled training data. Existing
methods to address this challenge can be mainly
divided into three categories: \Ni reducing the hypothesis space with prior knowledge \citep{triantafillou2017few,hu-etal-2018-shot,li-etal-2023-contrastive-learning}, \Nii optimizing the strategy for searching the best hypothesis in whole space \citep{ravi2016optimization,finn2017model,qin-etal-2023-lifelong}, and \Niii augmenting the few-shot data \citep{gao2020neural,qin-joty-2022-continual,ding-etal-2023-gpt}. More recently, LLMs have achieved promising performance on various few-shot tasks via in-context learning (ICL).

\subsection{In-context Learning (ICL)} 

By conditioning on a prompt that includes several demonstration examples and optionally a task description, a frozen LLM, by virtue of ICL, showcases impressive few-shot generalization \citep{brown2020language}. ICL has drawn a great deal of attention from the research community in recent days. \citet{chen-etal-2022-improving,min-etal-2022-metaicl,wei2023symbol} have explored ways to enhance the ICL capabilities of language models by either self-supervised or supervised training. In parallel, extensive analytical studies have been conducted to understand factors influencing the performance of ICL \citep{zhao21c,wei2022emergent,yoo-etal-2022-ground,min-etal-2022-rethinking,wei2023larger,yang2023exploring,zhang-etal-2024-study}, as well as to elucidate the underlying mechanisms that contribute to the success of ICL \citep{olsson2022context,xie2022an,pan2023context,li2023transformers,dai2023can,qin2024relevant}. Furthermore, there is a series of ongoing research dedicated to various aspects of ICL: \Ni demonstration designing strategies, including demonstration organization \citep{liu-etal-2022-makes,rubin-etal-2022-learning,wang2023large,qin2023context,wang-etal-2024-learning} and demonstration formatting \citep{wei2022chain,wang2022self,zhang2023automatic,zhou2023large}, \Nii multi-modal ICL \citep{huang2023language,wang2023context,wang2023neural,zhu2023minigpt}, and \Niii applications of ICL  \citep{ding2022gpt,meade2023using,zheng2023can,long-etal-2024-prompt}. 

\subsection{Alignment}


Existing work on alignment can be mainly divided into two parts based on the objectives: aligning with humans and aligning with teacher models. To align with humans, reinforcement learning from human feedback (RLHF) \citep{christiano2017deep,ouyang2022training} explores how human feedback can be used to train language models to better align with human preferences and values using reinforcement learning algorithms such as PPO \citep{schulman2017proximal}. {Several recent studies have introduced lightweight alternatives of PPO, including RRHF \citep{yuan2023rrhf}, DPO \citep{rafailov2023direct}, ReMax \citep{li2023remax}, IPO \citep{azar2024general} and KTO \citep{ethayarajh2024kto}.} Alignment with teacher models, also known as distillation \citep{hinton2015distilling}, aims to transfer the powerful capabilities of large teacher models to more efficient and compact student counterparts. It has emerged as a powerful solution to reduce the high computational demands and serving challenges posed by large models. {Current distillation methods typically train student models on generated outputs of teacher models \citep{hsieh-etal-2023-distilling,wang2022self,xu2023wizardlm} or to imitate teacher models' token-level probability distributions \citep{sanh2019distilbert,jiao-etal-2020-tinybert,agarwal2023gkd,huang2022context,gu2024minillm}, i.e., these approaches focus on aligning the output of student models with that of teachers. However, they pay little attention to the input demonstrations which also significantly influence the performance of ICL \citep{qin2023context}. In contrast to these methods, our proposed method (\ba) leverages the models' preferences for different in-context examples to achieve input preference alignment.}

\section{Methodology} \label{bialign_method}

\subsection{Problem Setting} \label{sec:problem_setting}

\begin{figure}[t]
   \centering
   \includegraphics[width=0.46\textwidth]{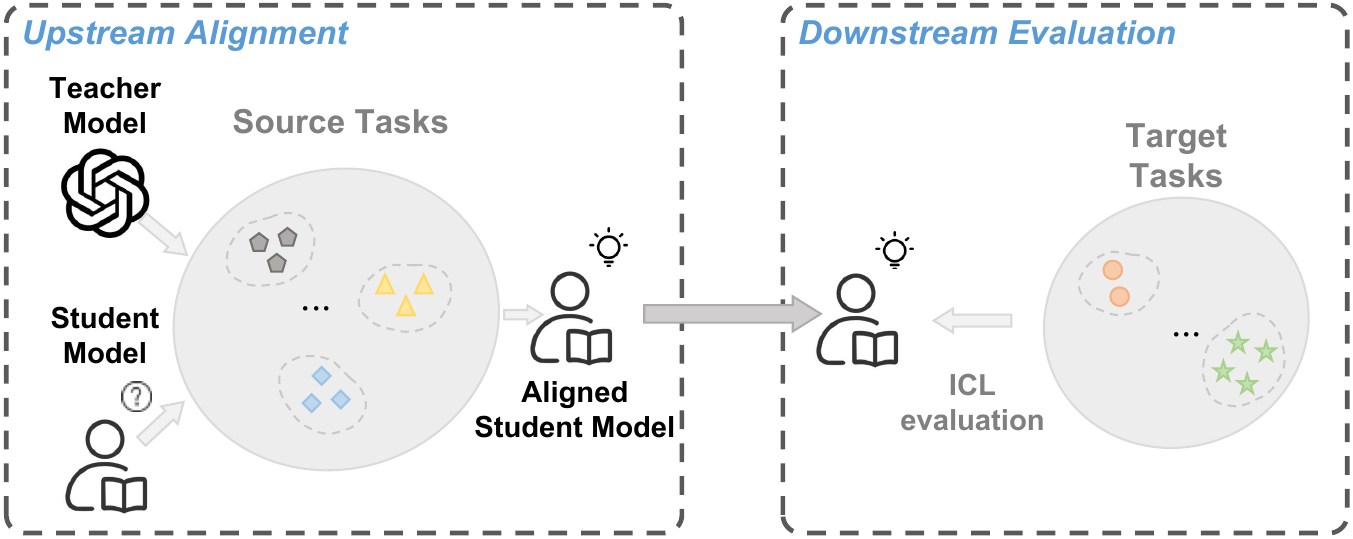}
  \caption{In the upstream ICL alignment stage, we align a student model with a teacher on the source tasks. Then in the downstream evaluation stage, we evaluate the ICL performance of the aligned student model on a held-out set of target tasks, which are different from the source tasks.
  }
  \label{fig:setting} 
\end{figure}

Given a training set $\mathcal{D}_{\text{train}}$ consisting of a set of source tasks $\mathcal{T}^\text{src}$, the goal of ICL alignment is to align the ICL ability of a student model $\mathrm{S}$ with that of a teacher model $\mathrm{T}$. 
Upon successful alignment, the model $\mathrm{S}$ is expected to show improved ICL ability on a held-out set of target tasks $\mathcal{T}^\text{tgt}$. We divide the whole process into two stages, as illustrated in Figure~\ref{fig:setting}.

\paragraph{$\bullet$ Upstream ICL alignment on $\mathcal{T}^\text{src}$:} In this alignment stage, the model has access to $\mathcal{T}^\text{src}$. We formalize samples in $\mathcal{D}_{\text{train}}$ in the $k$-shot ICL format $\{ \hat X_i = (x_1,y_1),...,(x_k,y_k),(\hat x_i,\hat y_i)\}$, where $(x_j,y_j), 1 \leq j \leq k$ denotes the $k$ demonstration examples and $(\hat x_i,\hat y_i)$ is the test sample. We concatenate these examples to form an ICL training sample $\hat X_i$. We then align the student model $\mathrm{S}$ with the teacher model $\mathrm{T}$ on this formatted ICL data.

\vspace{-0.5em}
\paragraph{$\bullet$ Downstream ICL evaluation on $\mathcal{T}^\text{tgt}$:} Following the upstream ICL alignment stage, we evaluate the ICL ability of the aligned model $\mathrm{S}^*$ on $\mathcal{T}^\text{tgt}$, where $\mathcal{T}^\text{tgt}$ has no overlap with $\mathcal{T}^\text{src}$. For every target task $\mathcal{T}_k$, we evaluate the model performance using both the default ICL demonstrations, as per convention, and their variants.

\begin{figure*}[t]
  \begin{center}
   \includegraphics[width=0.92\textwidth]{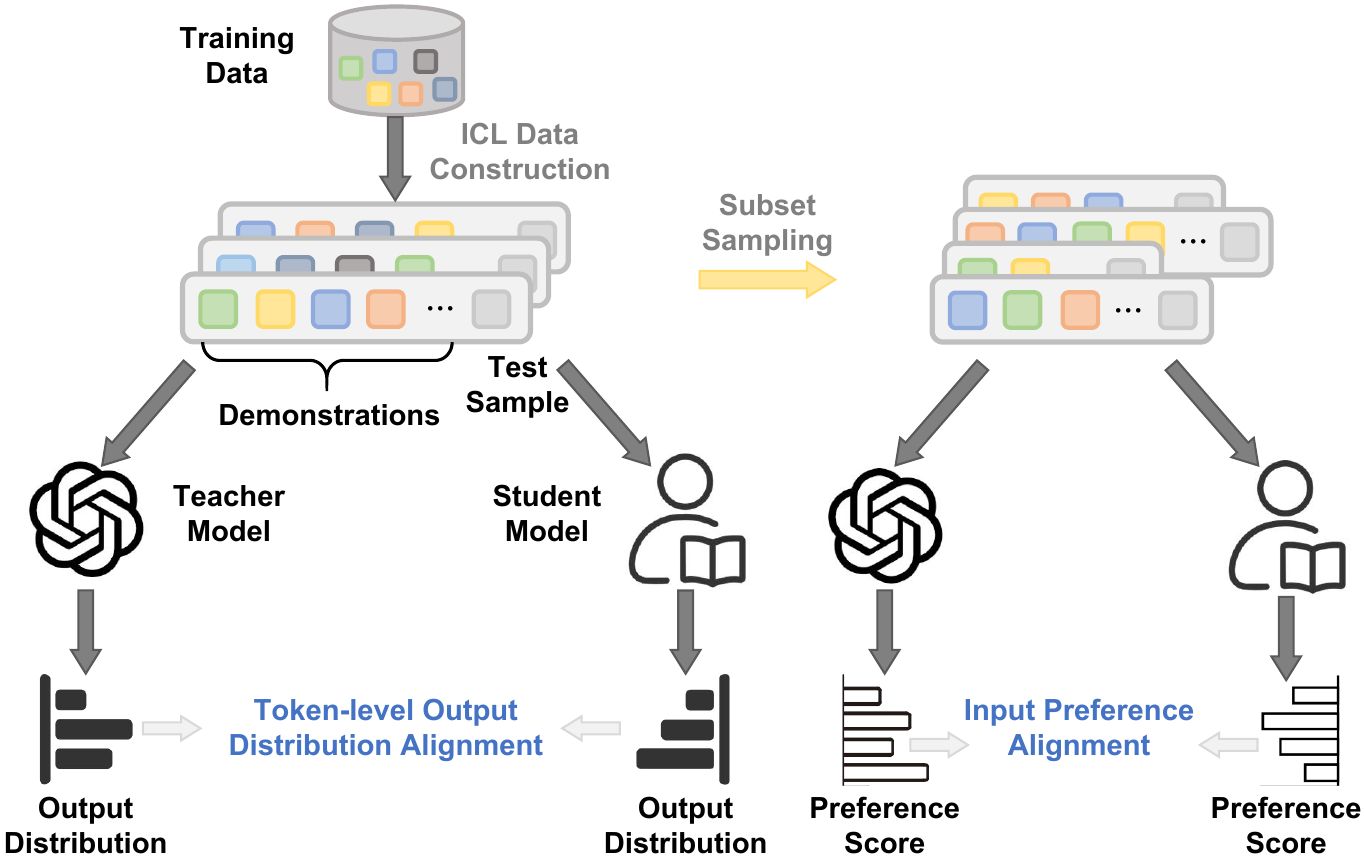}
  \end{center}
  \caption{Illustration of our Bidirectional Alignment (\ba) framework. It attains \textbf{\emph{token-level output distribution alignment}} by minimizing the KL divergence loss between the student and teacher models on the constructed ICL samples. Furthermore, after sampling several subsets from the set of all demonstrations, it optimizes a ranking loss for  \textbf{\emph{input preference alignment}} to align the student and teacher models' preferences for different demonstration examples. 
  }
  \label{fig:method}
\end{figure*}

\subsection{Bidirectional Alignment (\ba)} \label{sec:ba_method} 

Based on the finding that the performance of ICL is highly sensitive to the selection of demonstration examples \citep{zhao21c}, we propose Bidirectional Alignment (\ba) to fully leverage the models' preferences for different demonstration examples with the goal of improving the ICL ability of the student model. Our approach is illustrated in Figure~\ref{fig:method}.

\paragraph{Aligning Token-level Distributions}

Given the ICL training examples in the concatenated form $\{ \hat X_i = (x_1,y_1),...,(x_k,y_k),(\hat x_i,\hat y_i)\}$ as discussed above, to achieve \emph{token-level output distribution alignment} on $\hat X_i$, we minimize a KL divergence loss between the student model and teacher model for the \emph{whole} sequence instead of only $\hat{y}_i$ following \citet{gu2024minillm}.\footnote{Training on the whole sequence can ensure a large number of tokens in a batch, which is crucial to maintaining the basic in-weights capability \citep{chan2022data}.} More formally, 
\begin{equation}
\begin{aligned} \label{kl_loss}
\Scale[0.85]{\gL^{\text{KL}} =  \sum_{i=1}^{m} \sum_{j=1}^{t}  \KL (P_j (\gV | \hat X_i, \theta_{T}) || P_j (\gV | \hat X_i, \theta_{S}))}
\end{aligned}
\end{equation}
where $m$ is the number of ICL training samples in $\mathcal{D}_{\text{train}}$, $t$ is the number of tokens in $\hat X_i$, $\gV$ is the models' common vocabulary of tokens; $\theta_{T}$ and $\theta_{S}$ are the parameters of the teacher model and the student model, respectively. 

\paragraph{Aligning Preferences for Demonstrations}

Intuitively, for the student and teacher models to be well-aligned, the demonstrations preferred by the teacher model should also be preferred by the student, {i.e., to truly emulate the teacher model, the student needs to learn ``what to output'' as well as ``which input demonstrations should be preferred''} in order to generate high-quality outputs. This is similar in spirit to the scenario where a reward model is trained to align with preferences over model responses given by human experts \citep{ouyang2022training}. To this end, we introduce \emph{input preference alignment} to align the student and teacher models' preferences for different demonstrations.

For simplicity, let $\mathrm{R}_i = \{(x_1,y_1),...,(x_k,y_k)\}$ denote the $k$-shot  demonstrations in each ICL training sample $\hat X_i = (x_1,y_1),...,(x_k,y_k),(\hat x_i,\hat y_i)$. To rank the model's preferences for different demonstration examples, we first need to obtain a set $\mathcal{D}_{\text{rank}} = \{ \mathrm{R}_{ij}, (\hat x_i,\hat y_i) \}_{j=1}^{N}$, where $\mathrm{R}_{ij}$ is a subset of $\mathrm{R}_i$ and $N$ is the number of subsets considered for ranking. Modeling on the full subset space of $\mathrm{R}_i$ can be computationally prohibitive as it grows exponentially with $|\mathrm{R}_i|$. Therefore, we set $N \ll |\mathcal{P} (\mathrm{R}_i)|$, where  $\mathcal{P} (\mathrm{R}_i)$ is the power set of $\mathrm{R}_i$. \citet{zhao2024unveiling} highlights the impact of similar examples in the demonstrations. Building on this insight, we categorize all demonstrations in $\mathrm{R}_i$ into two groups, namely $\mathrm{G}_{sim}$ and $\mathrm{G}_{dissim}$, based on their similarity to the test example $(\hat x_i,\hat y_i)$ (see Appendix~\ref{sec:details_similarity} for details). Subsequently, we sample $N$ subsets from $\mathcal{P} (\mathrm{R}_i)$ with different numbers of similar examples.

We use both the student and teacher models to measure their preferences for each subset $\mathrm{R}_{ij}$, which we estimate using the prediction probability of $\hat y_i$ given $\mathrm{R}_{ij}$ and $\hat x_i$ as input:\footnote{Under the assumption that the prior $P(\mathrm{R}_{ij}|\hat x_i, \theta)$ is uniform, it is easy to show using the Bayes rule: $Q (\mathrm{R}_{ij}) \propto P(\mathrm{R}_{ij}| \hat y_i, \hat x_i,\theta) = \frac{P(\hat y_i| \mathrm{R}_{ij},  \hat x_i,\theta)  P(\mathrm{R}_{ij}| \hat x_i,\theta)}{\sum_j P(\hat y_i| \mathrm{R}_{ij},  \hat x_i,\theta)  P(\mathrm{R}_{ij}| \hat x_i,\theta)} $}
\begin{equation}
\begin{aligned} \label{perference}
\Scale[0.8]{Q^{\text{T}} (\mathrm{R}_{ij}) =  P ( \hat y_i |\mathrm{R}_{ij}, \hat x_i, \theta_{T}); Q^{\text{S}} (\mathrm{R}_{ij}) =  P ( \hat y_i |\mathrm{R}_{ij} , \hat x_i, \theta_{S})}
\end{aligned}
\end{equation}
where $Q^{\text{T}}$ and $Q^{\text{S}}$ are the preference scores of the teacher and student models, respectively. Intuitively, the more helpful the subset $\mathrm{R}_{ij}$ is for generating the target $\hat y_i$, the more the model prefers this subset.

To align the preferences of the student and teacher models for different subsets, we introduce a novel ranking loss:

\small
\begin{equation}
\begin{aligned} \label{rank_loss}
\gL^{\text{rank}} & =  \sum_{i=1}^{m} \sum_{\mathrm{R}^{+},\mathrm{R}^{-} \in \mathrm{R}_{i}^{\text{all}}} \max\{0, \\
& \underbrace{\frac{\log Q^{\text{S}} (\mathrm{R}^{-}) - \log Q^{\text{S}} (\mathrm{R}^{+})}{\mathop{\max}_{\mathrm{R}^{\prime} \in \mathrm{R}_{i}^{\text{all}}} \log Q^{\text{S}} (\mathrm{R}^{\prime}) - \mathop{\min}_{\mathrm{R}^{\prime} \in \mathrm{R}_{i}^{\text{all}}} \log Q^{\text{S}} (\mathrm{R}^{\prime})}}_{\text{\emph{Left}}} \\
& + \underbrace{\frac{1}{N - 1} (\text{rank}(Q^{\text{T}} (\mathrm{R}^{-})) - \text{rank}(Q^{\text{T}} (\mathrm{R}^{+})))}_{\text{\emph{Right}}} \}
\end{aligned}
\end{equation}
\normalsize
where $ \mathrm{R}_{i}^{\text{all}} = \{\mathrm{R}_{ij}\}_{j=1}^{N}$ contains all subsets sampled for the test example $(\hat x_i,\hat y_i)$, $(\mathrm{R}^{+}$, $\mathrm{R}^{-})$ refers to the pair of positive and negative subsets determined by the preference score of the teacher model (the subset with the higher preference score is considered as the positive one), and $\text{rank}()$ stands for the function that measures the relative ranking of subset scores which ranges from $1$ (most preferred) to $N$ (least preferred). The left part of $\gL^{\text{rank}}$ measures the difference in preference scores of the student model for the pair $(\mathrm{R}^{+},\mathrm{R}^{-})$ and the right part reflects the relative ranking difference between $\mathrm{R}^{+}$ and $\mathrm{R}^{-}$ (see more analysis of $\gL^{\text{rank}}$ in \Cref{sec:diff_rank}). Therefore,
$\gL^{\text{rank}}$ allows the student model to obtain more fine-grained supervision from the teacher model by \emph{matching the relative ranking} of their preference scores for different demonstration examples in ICL.

The overall loss that \ba\ optimizes for alignment is: $\gL = \gL^{\text{KL}} + \lambda \gL^{\text{rank}}$, where $\lambda$ is the weight of the ranking loss. Besides, we illustrate the whole learning process in Algorithm~\ref{alg:whole_process}.

\begin{algorithm}[t]
\caption{Learning process of \ba}
\textbf{Input:} ICL training set $\mathcal{D}_{\text{ICL}} = \{\hat X_i = (x_1,y_1),...,(x_k,y_k),(\hat x_i,\hat y_i) \}$, teacher model $\theta_{T}$, student model $\theta_{S}$, number of subsets $N$, weight of ranking loss $\lambda$
\begin{algorithmic}[1]
    \FOR{mini-batch $\mathcal{B}$ in $\mathcal{D}_{\text{ICL}}$}
    \STATE \textsc{Calculate} the KL divergence loss $\gL^{\text{KL}}$ on $\mathcal{B}$ using Equation~\ref{kl_loss}
    \FOR{$\hat X_i\ \in \mathcal{B}$    }
        \STATE \textsc{Sample} $N$ subsets $\{\mathrm{R}_{ij}\}_{j=1}^{N}$ for the test sample $(\hat x_i,\hat y_i)$
        \STATE \textsc{Measure} preferences $Q^{\text{T}}$ and $Q^{\text{S}}$ for $\{\mathrm{R}_{ij}\}_{j=1}^{N}$ using Equation~\ref{perference}
    \ENDFOR
    \STATE \textsc{Calculate} the ranking loss $\gL^{\text{rank}}$ on $\mathcal{B}$ using Equation~\ref{rank_loss}
    \STATE \textsc{Update} $\theta_{S}$ by back-propagating with  $\gL = \gL^{\text{KL}} + \lambda \gL^{\text{rank}}$
    \ENDFOR
\end{algorithmic}
\label{alg:whole_process} 
\end{algorithm}
\section{Experimental Setup} \label{exp_setup}

In this section, we first describe the tasks and datasets, and then introduce methods compared in our work.

\subsection{Tasks and Datasets} \label{sec:tasks_datasets}

In this work, we use CrossFit \citep{ye-etal-2021-crossfit}, a large and diverse collection of few-shot tasks covering various types including classification, question answering and generation, as the source tasks $\mathcal{T}^\text{src}$. For each task in CrossFit, we combine the original training and validation data as the new training data which is then randomly partitioned into a set of ICL samples with $4 \leq k \leq 10$ demonstration examples. For each ICL example, we sample $N = 4$ subsets from the set of all demonstrations for calculating the ranking loss. After the preprocessing, we obtain $12K$ ICL examples in total.

We evaluate the ICL performance of the aligned model on 5 target tasks spanning language understanding, symbolic reasoning, mathematical reasoning, logical reasoning, and coding: MMLU \citep{hendrycks2021measuring}, BBH \citep{suzgun2022challenging}, GSM8K \citep{cobbe2021training}, LogiQA \citep{liu2020logiqa} and HumanEval \citep{chen2021evaluating}. Note that there is no overlap between CrossFit and target tasks, and we obtain all outputs from the models using greedy decoding following \citet{xu2023lemur}. For each target task, we perform evaluations three times using different prompts and report the average results. Details of different target tasks and implementation are provided in Appendix~\ref{sec:detail_tgt_tasks} and \ref{sec:implementation_detail_ba}, respectively.

\begin{table*}[t]
    
    \centering
    \scalebox{0.92}{
    \begin{tabular}{
        lcccccc
        }
        \toprule
            \textbf{Method} &  MMLU & BBH & GSM8K & LogiQA & HumanEval & Average \\
        \midrule
& & & \textit{No Alignment Baselines} \\ 
        Vanilla &  $\text{45.4}_{\pm \text{0.6}}$ & $\text{39.5}_{\pm \text{0.5}}$ & $\text{15.2}_{\pm \text{0.3}}$ & $\text{30.3}_{\pm \text{0.4}}$ & $\text{14.6}_{\pm \text{0.4}}$ & $\text{29.0}_{\pm \text{0.3}}$ \\
        FT  & $\text{46.4}_{\pm \text{0.5}}$ & $\text{39.8}_{\pm \text{0.5}}$ &  $\text{15.6}_{\pm \text{0.4}}$ & $\text{31.7}_{\pm \text{0.3}}$ & $\text{14.2}_{\pm \text{0.4}}$ & $\text{29.5}_{\pm \text{0.4}}$ \\
        C-Pretrain & $\text{46.0}_{\pm \text{0.4}}$ & $\text{38.5}_{\pm \text{0.6}}$ & $\text{15.9}_{\pm \text{0.4}}$ & $\text{31.4}_{\pm \text{0.4}}$ & $\text{13.4}_{\pm \text{0.5}}$ & $\text{29.0}_{\pm \text{0.4}}$ \\
        \midrule
        Llama 2-13B Teacher\\
        \rowcolor{extremelightgray} \qquad Teacher & $\text{55.3}_{\pm \text{0.5}}$   & $\text{47.8}_{\pm \text{0.4}}$  & $\text{27.8}_{\pm \text{0.3}}$  & $\text{37.8}_{\pm \text{0.4}}$  & $\text{18.3}_{\pm \text{0.3}}$  &  $\text{37.4}_{\pm \text{0.3}}$\\ 
        \qquad Output-Align &  $\text{46.3}_{\pm \text{0.4}}$ & $\text{39.3}_{\pm \text{0.4}}$ & $\text{15.4}_{\pm \text{0.2}}$ & $\text{32.2}_{\pm \text{0.3}}$ & $\text{14.0}_{\pm \text{0.2}}$ & $\text{29.4}_{\pm \text{0.2}}$ \\
        \qquad \ba  &  $\textbf{47.5}_{\pm \text{0.4}}$ & $\textbf{41.0}_{\pm \text{0.3}}$ & $\textbf{16.8}_{\pm \text{0.3}}$ & $\textbf{33.9}_{\pm \text{0.4}}$ & $\textbf{15.6}_{\pm \text{0.4}}$ & $\textbf{31.0}_{\pm \text{0.3}}$ \\
        \midrule
        Llama 2-70B Teacher\\
        \rowcolor{extremelightgray}  \qquad Teacher & $\text{67.2}_{\pm \text{0.6}}$ & $\text{64.2}_{\pm \text{0.4}}$ & $\text{53.3}_{\pm \text{0.4}}$ & $\text{48.0}_{\pm \text{0.5}}$ & $\text{26.8}_{\pm \text{0.4}}$ & $\text{51.9}_{\pm \text{0.4}}$ \\
        \qquad Output-Align & $\text{47.1}_{\pm \text{0.5}}$ & $\text{39.8}_{\pm \text{0.4}}$ & $\text{16.4}_{\pm \text{0.3}}$ & $\text{33.2}_{\pm \text{0.3}}$ & $\text{14.6}_{\pm \text{0.4}}$ & $\text{30.2}_{\pm \text{0.3}}$ \\
        \qquad \ba  & $\textbf{49.5}_{\pm \text{0.3}}$ & $\textbf{43.2}_{\pm \text{0.5}}$ & $\textbf{18.3}_{\pm \text{0.4}}$ & $\textbf{35.7}_{\pm \text{0.4}}$ & $\textbf{16.6}_{\pm \text{0.3}}$ & $\textbf{32.7}_{\pm \text{0.3}}$ \\
        \bottomrule
    \end{tabular}
    }
    \caption{Performance ($\%$) of different methods on 5 target tasks. We use Llama 2-7B as a student and Llama 2-13B or 70B as a teacher model. The rows with ``Teacher" (\textcolor{gray}{grey}) indicate the corresponding teacher model's performance on the target tasks. \textbf{Bold} indicates the best result for Llama 2-7B (student). \ba\ is consistently better than all previous baselines. 
    }
    \label{tab:main-result}
\end{table*}

\subsection{Methods Compared} 
{We mainly experiment with Llama 2-7B \citep{touvron2023llama} as the student model and Llama 2-13B or 70B as the teacher model. For Llama 2-70B, we use the quantized version TheBloke/Llama-2-70B-GPTQ \citep{Llama-2-70B-GPTQ} due to resource constraints.} We compare \ba\ with the following methods:

\begin{itemize}[leftmargin=*]
    \item \textbf{Vanilla} simply evaluates the ICL performance of the student model on target tasks without any alignment, serving as the baseline for all other approaches. 
    \item \textbf{Fine-tuning (FT)} tunes the student model on the $12K$ ICL examples constructed from CrossFit using a multi-task learning scheme, which is indeed the meta-training in \citet{min-etal-2022-metaicl}. 
    \item \textbf{Continual Pretraining (C-Pretrain)} simply performs continual pretraining, \ie\ next token prediction for the whole sequence, of the student model on the $12K$ samples.
    \item \textbf{Output Alignment (Output-Align)} trains the student model to align token-level output distributions with the teacher model \citep{huang2022context,gu2024minillm}.
\end{itemize}

Additionally, we show the connection between \ba\ and In-Context Pretraining \citep{shi2024incontext} in \Cref{sec:connect_icp}, and discuss how \ba\ can be integrated with the latest ICL demonstration selection methods or reverse KL divergence in Appendix~\ref{sec:combine_icl_selection} and \ref{sec:combine_reverse_kl}.

\section{Results and Analysis} \label{result_and_analysis}

\subsection{Main Results}

Table~\ref{tab:main-result} shows the performance scores of different methods on all investigated target tasks. From the results, we can observe that

\noindent $\bullet~$ {Our proposed \ba\ consistently outperforms baseline approaches on all datasets with different sizes of teacher models, demonstrating its superiority. Simply pretraining the model on source tasks does not improve the average performance since there is no overlap between source and target tasks. While fine-tuning brings marginal improvement, token-level output distribution alignment with a stronger (70B) teacher model can achieve better performance. Thanks to incorporating input preference alignment (see \Cref{sec:computational_overhead} for analysis of computational overhead), \ba\ yields about 2.0$\%$ performance boost on average when using a 13B teacher model, and this gain is 3.7$\%$ for a 70B teacher. Besides, when examining the effects of scaling up the teacher model, the performance of \ba\ sees an improvement on all tasks.}

\begin{table}[t]
    
    \centering
    \scalebox{0.75}{
    \begin{tabular}{llcccc}
    \toprule
    \multicolumn{2}{l}{} & ASDiv & SVAMP & GSM8K & AQUA-RAT \\
    \midrule
    \multicolumn{2}{l}{Vanilla} & 46.6 & 41.2 & 15.2 & 24.4  \\
    \multicolumn{2}{l}{\ba} &  \textbf{49.4} & \textbf{43.5} & \textbf{16.8} & \textbf{27.2}   \\
    \midrule
    \multicolumn{2}{l}{Relative Gain} & 6.0 & 5.6 &  10.5 & 11.5  \\
    \bottomrule
    \end{tabular}
    }
    \caption{{Relative gain ($\%$) of \ba\ on  math reasoning tasks of varying difficulty levels.}}
    \label{tab:fine-grained}
\end{table}

\noindent $\bullet~$ In particular, \ba\ using a 13B teacher model achieves relative performance improvements of 11.9$\%$ on LogiQA and 10.5$\%$ on GSM8K compared with Vanilla, while using the 70B teacher, it achieves 17.8$\%$ on LogiQA and 20.4$\%$ on GSM8K. These results indicate that \ba\ can better improve the performance of tasks requiring more fine-grained reasoning; see appendix~\ref{sec:example_logiqa} for an example in LogiQA. This is because \ba\ allows the student model to obtain more fine-grained supervision from the teacher model by fully leveraging their preferences for different inputs.

To better support our claim, we further conduct experiments on four mathematical reasoning tasks ranging from low to high difficulty: ASDiv \citep{miao-etal-2020-diverse}, SVAMP \citep{svamp}, GSM8K \citep{cobbe2021training}, and AQUA-RAT \citep{aqua}. The comparison between \ba\ and Vanilla, as illustrated in Table~\ref{tab:fine-grained}, demonstrates that \ba\ is indeed more beneficial for more complex reasoning tasks.

\noindent $\bullet~$ Both fine-tuning and output alignment sometimes hurt the zero-shot learning capability of the model as shown by the performance on HumanEval. In contrast, \ba\ brings an average relative improvement of about 10.3$\%$ on HumanEval. We speculate that this is due to the subset sampling in input preference alignment, which helps the model generalize better to the unseen zero-shot setting.

\subsection{Analysis} \label{sec:more_analysis}

  \begin{table}[t]
  \centering
  \scalebox{0.90}{
    \begin{tabular}{lcc}
    \toprule
    \multirow{1}{*}{\textbf{Method}} & \multicolumn{1}{c}{7B}                    & \multicolumn{1}{c}{13B}                    \\
    \midrule
    Output-Align & 30.2  & 38.8 \\
    \ba & \textbf{32.7} & \textbf{40.9} \\
    \bottomrule
    \end{tabular}
    }
    \captionof{table}{{Average results ($\%$) of Output-Align and \ba\ with different sizes of student models (Llama 2-70B as the teacher).}}
    \label{table:large_student_model}
  \end{table}
  
  \begin{table}[t]
  \centering
  \scalebox{0.63}{
    \begin{tabular}{llccccc}
    \toprule
    \multicolumn{2}{l}{\textbf{Method}} & Vanilla & FT & C-Pretrain & Output-Align & \ba \\
    \cmidrule{3-7}
    \multicolumn{2}{l}{Llama 3-8B} &  60.4 & 61.0 & 60.5 & 61.7 & \textbf{63.9} \\
    \multicolumn{2}{l}{Phi-3-mini (3.8B)}  & 66.7 & 67.1 & 66.5 & 67.4 & \textbf{69.1} \\
    \bottomrule
    \end{tabular}
    }
    \captionof{table}{Average results ($\%$) across 5 tasks of all methods with two different backbones. We use Llama 3-70B as the teacher for Llama 3-8B and Phi-3-medium (14B) as the teacher for Phi-3-mini (3.8B).}
  \label{tab:diff_backbone}
\end{table}

\begin{table}[t]
    
    \centering
     \scalebox{0.9}{
    \begin{tabular}{lcc}
    \toprule
    \multirow{1}{*}{} & \multicolumn{1}{c}{Default}    & \multicolumn{1}{c}{Variant}     \\
    \midrule
    
    \ba  &   \textbf{31.0} & 30.5
    \\
    
    \bottomrule
    \end{tabular}
    }
    \caption{Average results ($\%$) of \ba\ with different ranking loss formulations.} 
    \label{table:different_loss}
\end{table}

\begin{figure}[t]
  \centering
    \includegraphics[width=0.45\textwidth]{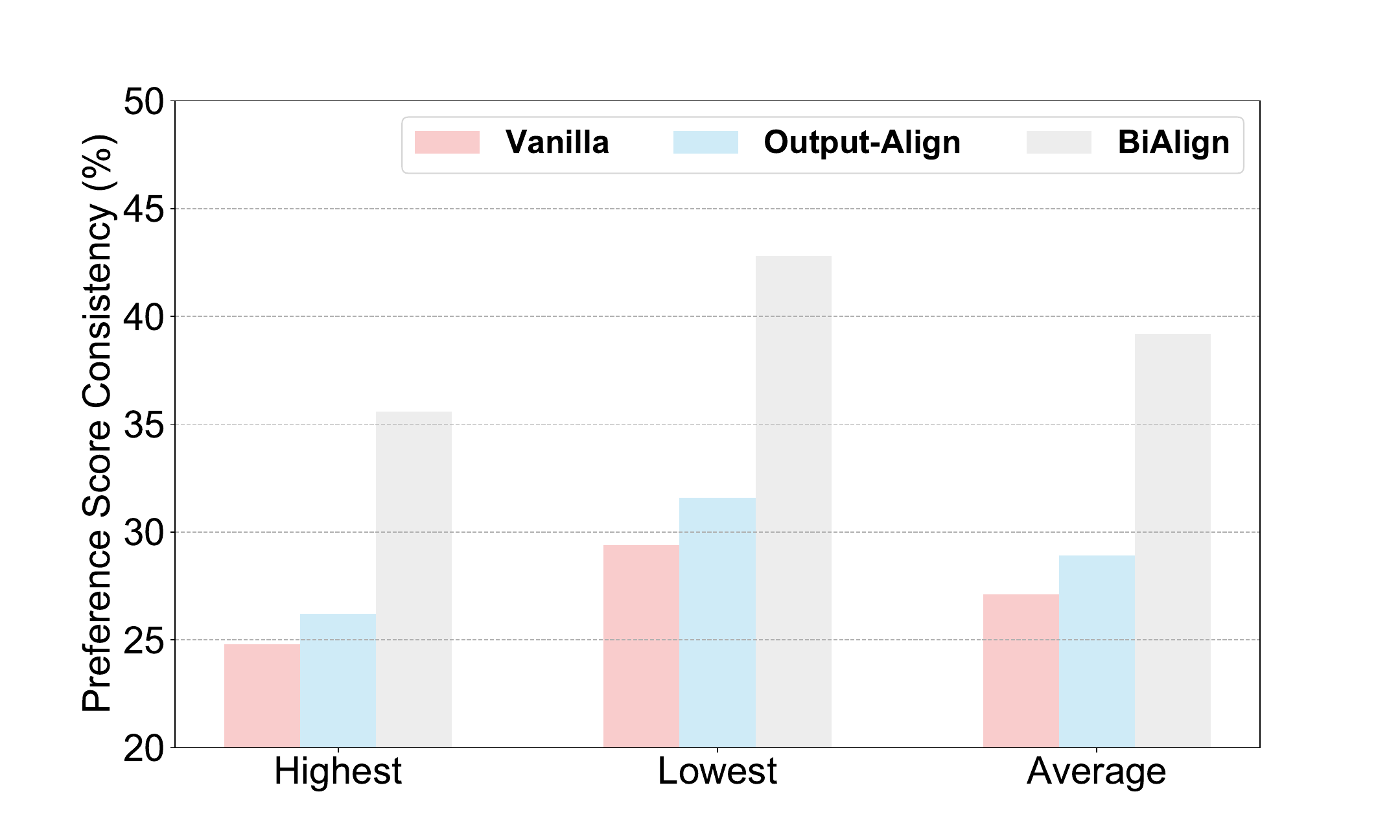}
 \captionof{figure}{Preference score consistency ($\%$) of different methods.}
  \label{fig:preference_consistency}
\end{figure}

\paragraph{Larger Student Model} We further experiment with a larger student model to verify the effectiveness of \ba. Specifically, we use Llama 2-13B as the student model and Llama 2-70B as the teacher model. We employ QLoRA \citep{dettmers2023qlora} to fine-tune the student model with consideration of computational resource limitations. The results averaged over the 5 tasks are reported in Table~\ref{table:large_student_model}, which demonstrate the consistent superiority of \ba\ across model scales.

\paragraph{Different Backbone Models} Our experiments and analysis so far use Llama 2 as the backbone model. To verify whether the performance gain of \ba\ is consistent across different backbone models, we extend the experiments to Llama 3 \citep{dubey2024llama} and Phi 3 \citep{abdin2024phi}. For Llama 3, we use the 8B model as the student and the 70B model as the teacher. For Phi 3, we use Phi-3-mini (3.8B) as the student and Phi-3-medium (14B) as the teacher. From the average results shown in Table~\ref{tab:diff_backbone}, we can see that \ba\ still outperforms all baseline approaches when using other language models as the backbone, showing its robustness to model types. In addition, we show the scalability of \ba\ across more model scales using Qwen-2.5 \citep{yang2024qwen2} in Appendix~\ref{sec:scalability_more_model_scales}. 

\paragraph{Comment on Training-time Computational Overhead} \label{sec:computational_overhead} Smaller models are a preferred choice for resource-constrained deployments, where the inference cost matters the most. \ba\ does not introduce any additional cost during inference. The additional computational overhead only occurs once during model training. To quantify the increase in computational overhead caused by the ranking loss, we use DeepSpeed Flops Profiler \citep{rasley2020deepspeed} to calculate the training FLOPs of Output-Align and \ba, which are 3.3$\times10^{17}$ and 7.6$\times10^{17}$ respectively (about 2.3 times). Therefore, we further design two experiments to compare \ba\ and Output-Align under the same training FLOPs: \Ni we combine the original ICL training examples with the sampled subset data and conduct Output-Align on the combined data (roughly the same FLOPs as \ba), which performs (29.5) similarly to the original Output-Align method (29.4), verifying the superiority of \ba; \Nii we reduce the training epochs of \ba\ from 4 to 2 (roughly the same FLOPs as Output-Align) and assess the final checkpoint. There is no significant performance degradation (from 31.0 to 30.8), which also demonstrates that \ba\ can outperform baselines under the same training FLOPs.

\paragraph{Different Ranking Loss Formulations} \label{sec:diff_rank} In the right part of Equation~\ref{rank_loss}, we employ the $\text{rank}()$ function to represent the relative ranking of the model's preference scores instead of relying on the scores themselves. This approach is grounded in the idea that the primary objective of input preference alignment is to match the rankings of the subset scores, rather than their specific values. By focusing on rankings, we can reduce the impact of potential variations in score magnitudes, allowing the model to prioritize the relative ranking of preferences. We further conduct experiments with an alternative ranking loss formulation that does not incorporate $\text{rank}()$, while maintaining all other implementation details. The average results reported in Table~\ref{table:different_loss} underscore the importance of using $\text{rank}()$ for alignment.

\paragraph{Connection with In-Context Pretraining} \label{sec:connect_icp} \Citet{shi2024incontext} propose In-Context Pretraining (ICP) which pretrains language models on a sequence of related documents. \ba\ mainly differs from it in the following two aspects: \Ni ICP focuses on the pretraining stage while \ba\ is specifically designed for more lightweight supervised fine-tuning. \Nii The objective of ICP is to design more effective pretraining data. In contrast, \ba\ leverages distillation to improve the capabilities of the student model. Therefore, \ba\ can be seamlessly integrated with ICP to further improve the ICL ability.

\paragraph{Effect of Demonstration Numbers} As mentioned in Section~\ref{sec:tasks_datasets}, each constructed ICL training sample contains $4 \leq k \leq 10$ demonstration examples, which could enhance the model's ability to generalize to different numbers of demonstrations. To investigate the effect of demonstration numbers in source tasks, we further conduct training on examples containing a fixed number $k \in \{5, 8, 10\}$ of demonstrations. The average results of the 5 target tasks are reported in Table~\ref{table:diff_demonstration}. We can see that training with a fixed number of demonstrations results in performance degradation to a certain degree, justifying our training set construction strategy.

\paragraph{Preference Score Consistency} {As illustrated in Section~\ref{sec:ba_method}, $\gL^{\text{rank}}$ enables the student model to match the relative ranking of the preference scores for different ICL demonstrations with that of the teacher model. To verify this, we report the \emph{preference score consistency} comparison between \ba\ and Output-Align in Figure~\ref{fig:preference_consistency}. Specifically, we randomly select $500$ examples from MMLU (see results on other datasets in Appendix~\ref{sec:preference_score_consistency_other_datasets}). For each example, we randomly sample $5$ subsets from the set of all demonstrations and obtain their preference scores using different models. The preference score consistency of different methods is then calculated as the proportion of the highest/lowest scoring subsets that are consistent between the corresponding student model and the teacher model. From the results, we can see that \ba\ can indeed achieve much higher preference score consistency than Output-Align, indicating the effectiveness of $\gL^{\text{rank}}$.}

  \begin{table}[t]
  \centering
  
  \scalebox{0.82}{
    \begin{tabular}{lcccc}
    \toprule
    \multirow{2}{*}{\textbf{Method}} & \multicolumn{4}{c}{\textbf{Demonstration number}} \\
    \cmidrule(lr){2-5}
    & Default ($4 \leq k \leq 10$) & 5 & 8 & 10 \\
    \midrule
    \ba & \textbf{31.0}  & 30.8  & 30.4  & 30.5  \\
    \bottomrule
    \end{tabular}
    }
    \captionof{table}{Average results ($\%$) of \ba\ with different $k$ (demonstration number) for constructed ICL training samples.}
  \label{table:diff_demonstration}
\end{table}

In addition, for interested readers, we show further justification of input preference alignment, more cross-task generalization experiments, the results with different subset sampling methods, different numbers of subsets and different source task selections, the analysis of KL divergence calculation, training steps and additional training data, the influence of ranking loss weight, the effect of contrastive pair selection, and a case study of model output in Appendix~\ref{sec:further_justification} $\sim$~\ref{sec:case_study_output}, respectively.
\section{Conclusion}

In this work, we have introduced Bidirectional Alignment (\ba) that can improve the ICL capabilities of student models by aligning the input preferences between student and teacher models in addition to aligning the token-level output distributions. Extensive experimental results and analysis show that \ba\ consistently outperforms previous baseline approaches.

\section*{Limitations}
As the first work on input preference alignment, one limitation of our paper is the additional computational overhead introduced by the ranking loss. A further improvement could be to explore more efficient input alignment methods to improve the ICL capabilities of student models.

\bibliography{anthology,custom}

\appendix

\section{Appendix}

\subsection{Details of Splitting Groups by Similarity} \label{sec:details_similarity}

We use Sentence-BERT \citep{reimers-gurevych-2019-sentence} to obtain contextual representations of the examples and employ cosine similarity to measure the similarity between these representations. Based on the similarity to the test example, we categorize all demonstrations into two groups, $\mathrm{G}_{sim}$ and $\mathrm{G}_{dissim}$, ensuring an approximately equal split between the two groups (\ie\ a 1:1 ratio).

\begin{table}[t]
    \centering
    \resizebox{0.95\linewidth}{!}{
    \begin{tabular}{
        l @{\hspace{2em}}
        cccccc
        }
        \toprule
            \textbf{} & CrossFit &  MMLU & BBH & GSM8K & LogiQA & HumanEval  \\
        \midrule
        \# \textbf{Samples} & 12K & 15K & 6.5K & 8.5K & 651 & 164  \\
        \# \textbf{Shot} & 4$\sim$10  & 5 &  3 & 8 & 5 & 0  \\
        \bottomrule
    \end{tabular}
    }
    \caption{Details of different datasets. \# refers to `the number of'. CrossFit \citep{ye-etal-2021-crossfit} is used to construct training data and other tasks are used for evaluation.
    }
\label{tab:information}    
\end{table}

\subsection{Details of Target Tasks} \label{sec:detail_tgt_tasks}

In this work, we construct the in-context learning evaluation suite based on the following datasets: 

\begin{itemize}[leftmargin=*,topsep=3pt,itemsep=3pt,parsep=0pt]

\item \textbf{MMLU}: The MMLU (Massive Multitask Language Understanding) benchmark \citep{hendrycks2021measuring} consists of 57 diverse tasks covering various fields like computer science, history and law, aiming to evaluate the knowledge obtained during pretraining. Following its original setup, we use $5$-shot ICL demonstrations for evaluation.
\item \textbf{BBH}: The BBH (BIG-Bench Hard) \citep{suzgun2022challenging} includes several types of reasoning tasks that are believed to be difficult for current language models. Following the guidelines in \citet{suzgun2022challenging}, we conduct the evaluation using $3$-shot ICL demonstration examples with chain-of-thought prompting \citep{cot_wei}.
\item \textbf{GSM8K}: The GSM8K \citep{cobbe2021training} dataset encompasses 8.5K grade school math word problems, aiming to evaluate the multi-step mathematical reasoning capabilities. We evaluate the ICL performance on it using $8$-shot in-context examples with chain-of-thought prompting.
\item \textbf{LogiQA}: LogiQA \citep{liu2020logiqa} is a logical reasoning benchmark sourced from logical examination papers intended for reading comprehension. Following \citet{jiao2023logicllm}, we conduct $5$-shot evaluation.
\item \textbf{HumanEval}: HumanEval \citep{chen2021evaluating}  consists of 164 programming challenges for evaluating coding capabilities. We follow the official zero-shot setting in \citet{chen2021evaluating} to verify whether bidirectional alignment hurts the zero-shot learning ability of models.
\end{itemize}

We summarize the details of all used datasets in Table~\ref{tab:information}.

\begin{table}[t]
  \centering
  \scalebox{0.85}{
    \begin{tabular}{lccc}
    \toprule
    \multirow{1}{*}{} & \multicolumn{1}{c}{KATE}  & \multicolumn{1}{c}{MMR}   & \multicolumn{1}{c}{IDS}                  \\
    \midrule
    $\text{Selection}_{\text{Vanilla}}$ & 18.1 & 17.4 & 19.2 \\
    $\text{Selection}_{\text{BiAlign}}$ & \textbf{20.2} & \textbf{19.3} & \textbf{20.8} \\
    \bottomrule
    \end{tabular}
    }
    \captionof{table}{Integration of \ba\ with ICL demonstration selection methods.} 
  \label{table:combination_icl_selection_methods}
\end{table}

\begin{table}[t]
  \centering
  \scalebox{0.85}{
    \begin{tabular}{lcc}
    \toprule
    \multirow{1}{*}{} & \multicolumn{1}{c}{Output-Align}  & \multicolumn{1}{c}{\ba} \\
    \midrule
    Llama 3-8B & 62.5 & \textbf{65.3} \\
    \bottomrule
    \end{tabular}
    }
    \captionof{table}{Integration of \ba\ with reverse KL divergence.} 
  \label{table:combination_reverse_KL_divergence}
\end{table}

\begin{table}[t]
  \centering
  \scalebox{0.85}{
    \begin{tabular}{lcccc}
    \toprule
    \multirow{1}{*}{} & \multicolumn{1}{c}{1.5B} & \multicolumn{1}{c}{3B}   & \multicolumn{1}{c}{7B}     & \multicolumn{1}{c}{14B}             \\
    \midrule
    Output-Align & 35.2 & 35.9 & 36.2 & 36.7 \\
    \ba & \textbf{36.9} & \textbf{38.0} & \textbf{38.8} & \textbf{40.1} \\
    \bottomrule
    \end{tabular}
    }
    \captionof{table}{Results for different teacher model sizes with a fixed 0.5B student (Qwen-2.5).} 
  \label{table:qwen_varying_teacher}
\end{table}

\begin{table}[t]
  \centering
  \scalebox{0.85}{
    \begin{tabular}{lcccc}
    \toprule
    \multirow{1}{*}{} & \multicolumn{1}{c}{0.5B}  & \multicolumn{1}{c}{1.5B} & \multicolumn{1}{c}{3B}   & \multicolumn{1}{c}{7B}                 \\
    \midrule
    Output-Align & 36.7 & 51.4 & 59.6 & 70.9 \\
    \ba & \textbf{40.1} & \textbf{54.3} & \textbf{62.7} & \textbf{73.4} \\
    \bottomrule
    \end{tabular}
    }
    \captionof{table}{Results for different student model sizes with a fixed 14B teacher (Qwen-2.5).} 
  \label{table:qwen_varying_student}
\end{table}

\begin{table}[t]
  \centering
  
  \scalebox{0.85}{
    \begin{tabular}{lccc}
    \toprule
    \multirow{1}{*}{} & \multicolumn{1}{c}{Vanilla}                    & \multicolumn{1}{c}{Output-Align}   & \multicolumn{1}{c}{\ba}                  \\
    \midrule
    BBH & 31.4 & 33.8 & \textbf{45.3} \\
    GSM8K & 24.7 & 28.4 & \textbf{38.6} \\
    LogiQA & 29.1 & 32.3 & \textbf{44.7} \\
    \bottomrule
    \end{tabular}
    }
    \captionof{table}{Average preference score consistency ($\%$) comparison between different methods.} 
  \label{table:average_preference_score_c}
\end{table}

\begin{table}[t]
\centering
\setlength\tabcolsep{3pt}
\scalebox{0.9}{
\begin{tabular}{lcc}
\toprule
\multirow{1}{*}{} & \multicolumn{1}{c}{Output-Align}    & \multicolumn{1}{c}{\ba}     \\
\midrule
BBH  &  40.2 & \textbf{43.3}
\\

\bottomrule
\end{tabular}
}
\caption{Performance on BBH for models trained on MMLU.} 
\label{table:eval_bbh_train_mmlu}
\end{table}

\begin{table}[t]
\centering
\setlength\tabcolsep{3pt}
\scalebox{0.9}{
\begin{tabular}{lcc}
\toprule
\multirow{1}{*}{} & \multicolumn{1}{c}{Default}    & \multicolumn{1}{c}{Variant}     \\
\midrule

\ba  &   \textbf{31.0} & 30.3
\\

\bottomrule
\end{tabular}
}
\caption{Comparison between different subset sampling methods.} 
\label{table:diff_subset_sampling}
\end{table}

\begin{table}[t]
  \centering
  
  \scalebox{0.9}{
    \begin{tabular}{lcccc}
    \toprule
    \multirow{2}{*}{\textbf{Method}} & \multicolumn{4}{c}{\textbf{Subset number}} \\
    \cmidrule(lr){2-5}
    & 3 & 4 & 5 & 6 \\
    \midrule
    \ba & 30.7  & 31.0  & 30.8  &  \textbf{31.1}  \\
    \bottomrule
    \end{tabular}
    }
    \captionof{table}{Average performance ($\%$) of \ba\ with different numbers of subsets $N$.}
  \label{table:diff_subset}
\end{table}

 \begin{figure}[t]
  \centering
  \includegraphics[width=0.44\textwidth]{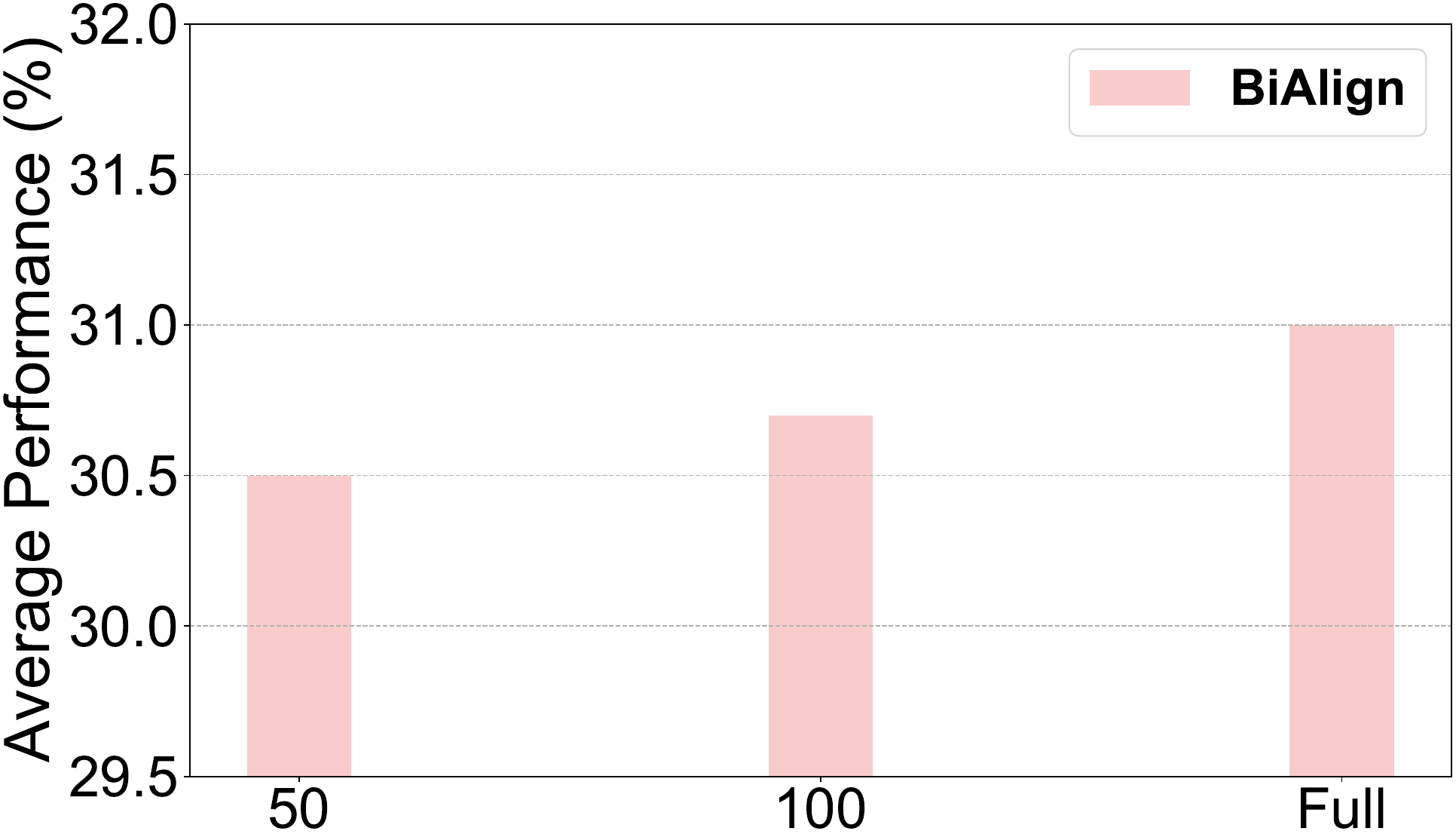}
    \captionof{figure}{Average performance ($\%$) of \ba\ with different numbers of source tasks.}
    \label{fig:diff_source}
\end{figure}

\begin{table}[t]
 \centering
 
  \scalebox{0.9}{
   \begin{tabular}{lcccc}
    \toprule
    \multirow{2}{*}{\textbf{Method}} & \multicolumn{2}{c}{\textbf{Type}} \\
    \cmidrule(lr){2-3}
    & Whole Sequence & Label Only  \\
    \midrule
    \ba & \textbf{31.0}  &  30.8 \\
    \bottomrule
    \end{tabular}
    }
    \captionof{table}{Average performance ($\%$) of \ba\ using different types of KL divergence calculation methods.} 
  \label{table:whole_seq_onlylabel}
  \end{table}
  
  \begin{table}[t]
  \centering
  
  \scalebox{0.85}{
    \begin{tabular}{lccc}
    \toprule
    \multirow{1}{*}{\textbf{Method}} & \multicolumn{1}{c}{25$\%$}                    & \multicolumn{1}{c}{50$\%$}   & \multicolumn{1}{c}{100$\%$}                  \\
    \midrule
    Output-Align & 29.1  &  29.3 & 29.4 \\
    \ba & \textbf{30.3} & \textbf{30.8} & \textbf{31.0} \\
    \bottomrule
    \end{tabular}
    }
    \captionof{table}{Comparison between \ba\ and Output-Align at different proportions of training steps.} 
  \label{table:diff_prop_training_steps}
\end{table}

\subsection{Implementation Details} \label{sec:implementation_detail_ba}

Our methods are implemented with the PyTorch and Transformers library \citep{wolf-etal-2020-transformers}. Model training is conducted utilizing DeepSpeed \citep{rasley2020deepspeed,rajbhandari2020zero} on 4 NVIDIA A100 GPUs. During the training phase, we set the learning rate to $1\mathrm{e}{-6}$ and the batch size to 64. The weight $\lambda$ for the ranking loss is set to 1.0. For evaluation, we train the student model on the constructed ICL data for $4$ epochs and assess the final checkpoint.

\subsection{Combination with ICL Demonstration Selection Methods} \label{sec:combine_icl_selection}

\ba\ is complementary to ICL demonstration selection methods and can be seamlessly integrated with them to further improve ICL performance. To validate this, we investigate three demonstration selection methods: KATE \citep{liu-etal-2022-makes}, MMR \citep{ye-etal-2023-complementary}, and IDS \citep{qin2023context}. For each method, we evaluate the following two variants: selecting demonstrations and performing ICL using the vanilla model ($\text{Selection}_{\text{Vanilla}}$), and selecting demonstrations and performing ICL using the model after \ba\ ($\text{Selection}_{\text{BiAlign}}$). We conduct experiments on GSM8K and report the results in Table~\ref{table:combination_icl_selection_methods}, demonstrating that \ba\ consistently boosts performance across all three selection methods. Furthermore, \ba\ (18.3) surpasses both KATE (18.1) and MMR (17.4), highlighting its superiority over several ICL demonstration selection approaches.

\subsection{Combination with Reverse KL Divergence} \label{sec:combine_reverse_kl}

\citet{gu2024minillm} reveals that reverse KL divergence is more suitable for knowledge distillation in generative LLMs, as it helps prevent the student model from overestimating low-probability regions of the teacher's distribution. Building on this insight, we investigate the integration of \ba\ with reverse KL divergence. Specifically, we replace the forward KL divergence in both Output-Align and \ba\ with reverse KL divergence and conduct experiments using Llama 3-70B as the teacher model and Llama 3-8B as the student model. As shown in Table~\ref{table:combination_reverse_KL_divergence}, \ba\ continues to significantly outperform Output-Align with reverse KL divergence, further demonstrating its effectiveness.

\subsection{Scalability to More Model Scales} \label{sec:scalability_more_model_scales}

We investigate the scalability of \ba\ across different model scales using Qwen-2.5 \citep{yang2024qwen2}. Specifically, we conduct experiments on five model scales: 0.5B, 1.5B, 3B, 7B, and 14B. Our evaluation follows two settings: \Ni \emph{Varying teacher model sizes}: We fix the student model at 0.5B and experiment with teacher models ranging from 1.5B to 14B. \Nii \emph{Varying student model sizes}: We fix the teacher model at 14B and test student models ranging from 0.5B to 7B. The results for both settings are presented in Table~\ref{table:qwen_varying_teacher} and \ref{table:qwen_varying_student}, respectively. We can see that:
\begin{itemize}[leftmargin=*]
    \item \ba\ consistently outperforms Output-Align with different sizes of teacher models.
    \item \ba\ \emph{benefits more from increasing the size of the teacher model} compared to Output-Align.
    \item \ba\ is both applicable and robust across student models of different sizes.
\end{itemize}

\subsection{Average Preference Score Consistency} \label{sec:preference_score_consistency_other_datasets}

We report the average preference score consistency ($\%$) comparison between different methods on the other three datasets (BBH, GSM8K and LogiQA) in Table~\ref{table:average_preference_score_c}. From the results, we can see that \ba\ consistently outperforms Output-Align across all datasets.

\subsection{Further Justification of Input Preference Alignment} \label{sec:further_justification}

We outline the justification for input preference alignment from the following perspectives.

\Ni \emph{Impact of ICL Demonstrations on Model Performance.} ICL demonstrations have been shown to have a significant impact on the performance of ICL \citep{liu-etal-2022-makes,qin2023context}. Selecting different sets of demonstration examples can yield performance ranging from almost random to better than state-of-the-art fine-tuned models, indicating that the model has different preferences for different inputs. For the student and teacher models to be well-aligned, the demonstrations preferred by the teacher model should also be preferred by the student, \ie\ \textbf{to truly emulate the teacher model, the student model needs to learn ``what to output'' as well as ``which input examples should be preferred''}. This is closely related to preference learning in RLHF, where the reward model learns "which outputs should be preferred". After learning, a well-trained reward model can rank model responses with expertise comparable to humans. To this end, we introduce input preference alignment to align the student and teacher models' preferences for different demonstrations.

\Nii \emph{Explanatory Mechanisms of ICL.} Another perspective supporting input preference alignment stems from the way LLMs process and prioritize information during ICL. \citet{kossen2024incontext} discover that LLMs do not treat all available information equally; instead, they exhibit a natural tendency to prioritize information closer to the query. This selective attention mechanism suggests that LLMs inherently favor contextually relevant details over more distant or less relevant ones. Building on this insight, our proposed input preference alignment ensures that \textbf{the student model learns to replicate the teacher model's information prioritization strategy}. By aligning the student’s input selection process with that of the teacher, we make the learning process more effective. This joint alignment ultimately enables the student model to utilize information in a manner consistent with the teacher model's intrinsic preferences, thereby improving its overall ICL performance.

\subsection{More Cross-Task Generalization Experiments} \label{more_cross_task}

To further verify the cross-task generalization ability of \ba, we train the model on MMLU and evaluate it on BBH. Specifically, we use Llama 2-7B as the student model and Llama 2-70B as the teacher model. The results reported in Table~\ref{table:eval_bbh_train_mmlu} highlight the superiority of \ba\ over Output-Align.

\subsection{Different Subset Sampling Methods} \label{sec:diff_subset_sampling_method}

To investigate the influence of subset sampling methods, we replace the original method with `Randomly sample N subsets' which does not consider similarity. The comparison between the two methods is shown in Table~\ref{table:diff_subset_sampling}. We can observe a noticeable performance drop, highlighting the crucial role of incorporating example similarity in the sampling process.

\subsection{Different Numbers of Subsets} \label{sec:diff_num_subset}

While we use $N = 4$ subsets for calculating the ranking loss, we also evaluate the effectiveness of \ba\ with different $N$. Specifically, we conduct controlled experiments with $\{3, 5, 6\}$ subsets and report the average results of the 5 target tasks in Table~\ref{table:diff_subset}. We can observe that increasing the number of subsets does not always improve performance. \ba\ achieves the best performance (31.1) with 6 subsets and the performance with 4 subsets (31.0) is comparable. Besides, all variants consistently outperform baseline methods in Table~\ref{tab:main-result}, demonstrating the effectiveness of our designed input preference alignment.

\subsection{Different Source Task Selections} \label{sec:diff_source_task}

We hypothesize that the diversity of source tasks has a considerable influence on target task performance. To verify this, we study the effect of the number of source tasks by conducting controlled experiments on $\{50,100\}$ randomly selected source tasks. From the results in Figure~\ref {fig:diff_source}, we can observe that the performance of \ba\ keeps improving as the number of source tasks increases, indicating the importance of source task diversity.

\subsection{Whole Sequence vs. Label Only}  \label{sec:wholeseq_labelpnly}

To maintain the basic in-weights capability of the student model, we minimize the KL divergence loss for the whole sequence instead of only the label following \citet{gu2024minillm}. In Table~\ref{table:whole_seq_onlylabel}, we show the performance comparison between using the whole sequence and using only the label. We can see that using the whole sequence also results in slightly better average performance.

\subsection{Different Proportions of Training Steps}

Table~\ref{table:diff_prop_training_steps} reports the performance comparison between \ba\ and Output-Align at different proportions (roughly 25$\%$, 50$\%$, and 100$\%$) of training steps. We can observe that \ba\ consistently outperforms Output-Align at different steps.

\subsection{Additional Training Data} 

The analysis in Section~\ref{sec:computational_overhead} shows that conducting Output-Align on the combination of the original ICL training examples and the sampled subset data achieves similar performance to the original Output-Align method. We further experiment with the fine-tuning approach. However, the performance becomes even worse (from 29.5 to 29.3), once again demonstrating that simply increasing training data does not necessarily lead to better performance.

\subsection{Ranking Loss Weights} \label{sec:ranking_loss}

To further investigate the influence of the ranking loss $\gL^{\text{rank}}$ (Equation~\ref{rank_loss}), we conduct experiments with different weights $\lambda$ and report the results in Table~\ref {table:diff_rank_loss}. All variants except the variant with $\lambda = 5.0$ (too large) outperform baseline approaches by a large margin, which demonstrates the superiority of $\gL^{\text{rank}}$.

\subsection{Contrastive Pair Selection} \label{sec:contrastive_pair_selection}

While we use all $C(N,2)$ ($N=4$ is the number of subsets) pairs of positive and negative subsets for input preference alignment, we also investigate the effect of contrastive pair selection. Specifically, we conduct controlled experiments on $\{3, 4, 5\}$ randomly selected contrastive pairs and report the average results in Table~\ref{table:contrastive_pair}. The best performance is achieved using all pairs, justifying our selection strategy.

\begin{table}[t]
  \centering
 
  \scalebox{0.85}{
   \begin{tabular}{lccccc}
    \toprule
    \multirow{2}{*}{\textbf{Method}} & \multicolumn{4}{c}{$\mathbf{\lambda}$} \\
    \cmidrule(lr){2-6}
    & 0.2 & 0.5 & 1.0 & 2.0 & 5.0 \\
    \midrule
    \ba & 30.8  & \textbf{31.2}  & 31.0  & 30.9 & 29.9 \\
    \bottomrule
    \end{tabular}
    }
    \captionof{table}{Average performance ($\%$) of \ba\ with different $\lambda$ for the ranking loss $\gL^{\text{rank}}$.} 
  \label{table:diff_rank_loss}
  \end{table}
  
  \begin{table}[t]
  \centering
  
  \scalebox{0.85}{
    \begin{tabular}{lcccc}
    \toprule
    \multirow{2}{*}{\textbf{Method}} & \multicolumn{4}{c}{\textbf{Pair number}} \\
    \cmidrule(lr){2-5}
    & 3 & 4 & 5 & All (6) \\
    \midrule
    \ba &  30.2 & 30.8  & 30.7  &  \textbf{31.0}  \\
    \bottomrule
    \end{tabular}
    }
    \captionof{table}{Average results ($\%$) of \ba\ with different numbers of contrastive pairs.} 
  \label{table:contrastive_pair}
  \end{table}

\subsection{Case Study} \label{sec:case_study_output}

We select GSM8K as a representative task and show several examples of output in Table~\ref{tab:diff_output}. Compared with Vallina, \ba\ is able to generate more precise and fine-grained reasoning paths, \eg\ \ba\ can successfully understand the meaning of `\emph{pair}' and generate the rationale `\emph{10 pairs of socks is 10 * 2 = 20 socks}' while Vallina fails to do so.

\begin{table*}[t]
\centering
\footnotesize
\begin{tabular}{p{0.95\textwidth}}
\toprule
Answer the following questions. \newline
\emph{Few-shot demonstrations} \\
\textbf{Question}: Lindsay is doing the laundry, and thinks she has missed some socks. There are 50 socks that need washing. If she washes 10 pairs of socks and 15 loose socks, how many socks has Lindsay missed? \\
\textbf{Answer}: \\
\midrule
\textbf{Vanilla}: Lindsay started with 50 socks. She washed 10 pairs of socks and 15 loose socks. \textcolor{red}{So she had 50 - 10 - 15 = 25 socks left. So the answer is 25.} \ngmark \\
\midrule
\textbf{\ba}: Lindsay started with 50 socks. She washed 10 pairs of socks and 15 loose socks. \textcolor{applegreen}{10 pairs of socks is 10 * 2 = 20 socks. 15 loose socks is 15 * 1 = 15 socks. So she missed 50 - 20 - 15 = 15 socks. So the answer is 15.} \okmark \\
\bottomrule
\end{tabular}
\caption{Output examples of Vanilla and \ba\ on GSM8K. The upper part is the input to LLMs, including few-shot demonstrations, and the lower part is the generated response. We color correct outputs in \textcolor{applegreen}{green} and wrong outputs in \textcolor{red}{red}.}
\label{tab:diff_output}
\end{table*}

\begin{table*}[t]
\centering
\footnotesize
\begin{tabular}{p{0.95\textwidth}}
\toprule
\textbf{Context}: There is no doubt that minors should be prohibited from smoking. However, we cannot explicitly ban the use of automatic cigarette vending machines in order to prevent minors from smoking. This ban is just like setting up roadblocks on the road to prohibit driving without a license. These roadblocks naturally prohibit driving without a license, but also block more than 99$\%$ of licensed drivers.\\
\textbf{Question}: In order to evaluate the above argument, which of the following questions is the most important? \\
\textbf{Options}: \\
A: Does the proportion of underage smokers in the total number of smokers exceed 1$\%$? \\
B: How much inconvenience does the ban on the use of automatic vending machines bring to adult cigarette buyers? \\
C: Whether the proportion of unlicensed drivers in the total number of drivers really does not exceed 1$\%$? \\
D: Is the harm of minor smoking really as serious as the public thinks? \\
\midrule
\textbf{Reasoning path}: 

To evaluate the argument effectively, the focus should be on understanding the impact and justification of the proposed ban on cigarette vending machines, especially in the context of preventing minors from smoking. The argument draws a parallel between the proposed ban and the hypothetical scenario of setting up roadblocks to prevent driving without a license, suggesting that while the measure may target a minority (in this case, underage smokers or unlicensed drivers), it disproportionately inconveniences the majority (licensed drivers or adult smokers). \newline

\emph{The most important question to evaluate the argument is: B: How much inconvenience does the ban on the use of automatic vending machines bring to adult cigarette buyers?} \newline

This option directly addresses the central issue of the argument—the balance between the inconvenience caused to the majority by the ban and the intended benefit of preventing a minority (minors) from engaging in harmful behavior (smoking). It's critical to assess whether the inconvenience to adult smokers (who are the majority of cigarette consumers) is justified by the potential benefits of reducing underage smoking. The comparison with roadblocks for unlicensed drivers emphasizes the argument's concern with the proportionality and fairness of broad preventive measures.\newline

Option A concerns the proportion of underage smokers but doesn't directly address the balance of convenience versus benefit. Option C relates to the analogy used but does not directly help in evaluating the effectiveness and appropriateness of the vending machine ban. Option D questions the severity of the problem (minor smoking) but does not specifically help in assessing the argument's concern about the consequences of the ban on the majority.\newline

Therefore, understanding the extent of inconvenience to adult smokers is crucial in determining whether the proposed solution is proportionate and justified, making Option B the most relevant and important question for evaluating the argument.\\
\bottomrule
\end{tabular}
\caption{Reasoning path for an example in LogiQA.}
\label{tab:logiqa_example}
\end{table*}

\subsection{Example in LogiQA} \label{sec:example_logiqa}

We show the reasoning path required to solve a sample in LogiQA in Table~\ref{tab:logiqa_example}, which is quite fine-grained.

\end{document}